\documentclass{article}
\usepackage{graphicx}
\usepackage{graphicx}
\usepackage{caption}
\usepackage{arydshln}
\usepackage{multirow}
\usepackage{natbib}
\usepackage{enumitem}
\usepackage{subcaption}
\usepackage{wrapfig}
\usepackage{float}
\usepackage{amsmath}

\PassOptionsToPackage{numbers, compress}{natbib}
\usepackage[preprint]{neurips_2024}

\usepackage[utf8]{inputenc} % allow utf-8 input
\usepackage[T1]{fontenc}    % use 8-bit T1 fonts
\usepackage{hyperref}       % hyperlinks
\usepackage{url}            % simple URL typesetting
\usepackage{booktabs}       % professional-quality tables
\usepackage{amsfonts}       % blackboard math symbols
\usepackage{nicefrac}       % compact symbols for 1/2, etc.
\usepackage{microtype}      % microtypography
\usepackage{xcolor}         % colors

\title{Twin Network Augmentation: A Novel Training Strategy for Improved Spiking Neural Networks and Efficient Weight Quantization}

\author{%
  Lucas Deckers \\
IDLab, University of Antwerp - imec\\
Sint-Pietersvliet 7\\
Antwerp, Belgium\\
  \texttt{lucas.deckers@uantwerpen.be} 
  \And
  Benjamin Vandersmissen \\
  IDLab, University of Antwerp - imec\\
Sint-Pietersvliet 7\\
Antwerp, Belgium\\
  \texttt{benjamin.vandersmissen@uantwerpen.be} 
  \AND
  Ing Jyh Tsang \\
  imec\\
Kapeldreef 75\\ 
3001 Leuven, Belgium\\
  \texttt{inton.tsang@imec.be} 
  \And
  Werner Van Leekwijck \\
  imec\\
Kapeldreef 75\\ 
3001 Leuven, Belgium\\
  \texttt{werner.vanleekwijck@imec.be} 
  \And
  Steven Latr\'e \\
  IDLab, University of Antwerp - imec\\
Sint-Pietersvliet 7\\
Antwerp, Belgium\\
  \texttt{steven.latre@uantwerpen.be} 
}

\begin{document}

\maketitle

\begin{abstract}

%%% Leave the Abstract empty if your article does not require one, please see the Summary Table for full details.
The proliferation of Artificial Neural Networks (ANNs) has led to increased energy consumption, raising concerns about their sustainability. Spiking Neural Networks (SNNs), which are inspired by biological neural systems and operate using sparse, event-driven spikes to communicate information between neurons, offer a potential solution due to their lower energy requirements. An alternative technique for reducing a neural network's footprint is quantization, which compresses weight representations to decrease memory usage and energy consumption.
In this study, we present Twin Network Augmentation (TNA), a novel training framework aimed at improving the performance of SNNs while also facilitating an enhanced compression through low-precision quantization of weights. TNA involves co-training an SNN with a twin network, optimizing both networks to minimize their cross-entropy losses and the mean squared error between their output logits. We demonstrate that TNA significantly enhances classification performance across various vision datasets and in addition is particularly effective when applied when reducing SNNs to ternary weight precision. Notably, during inference , only the ternary SNN is retained, significantly reducing the network in number of neurons, connectivity and weight size representation. Our results show that TNA outperforms traditional knowledge distillation methods and achieves state-of-the-art performance for the evaluated network architecture on benchmark datasets, including CIFAR-10, CIFAR-100, and CIFAR-10-DVS. This paper underscores the effectiveness of TNA in bridging the performance gap between SNNs and ANNs and suggests further exploration into the application of TNA in different network architectures and datasets.

\end{abstract}

\section{Introduction}

In recent years, there has been a significant increase in the utilization of Artificial Neural Networks (ANNs) across a wide range of application domains. A growing concern associated with this widespread utilization is the resulting increase in energy consumption. \cite{Iea}. A promising approach for alleviating the impacts of increasing demands in artificial intelligence is the implementation of spiking neural networks (SNNs). \cite{maass1997networks}. SNN are a type of biologically inspired networks that use binary (0 and 1) events, called spikes, to transmit information from one layer to another. In contrast to ANNs, the sparse, event-driven asynchronous processing of spikes \cite{bouvier2019spiking} in SNNs enables their deployment on specialized low-power neuromorphic hardware, such as Loihi \cite{orchard2021efficient} or SENeCA \cite{yousefzadeh2022seneca}. Recently \cite{shrestha2024efficient}, an SNN implementation was shown to gain up to three orders of magnitude in energy efficiency, latency, and even throughput on a video/audio processing task. An additional approach to enhancing the energy efficiency of neural networks involves network compression through weight quantization. \cite{courbariaux2015binaryconnect}. Integrating low-resolution weights with SNN facilitates efficient asynchronous information processing within on-chip memory systems.\cite{nguyen2022low}.

While SNNs demonstrate clear advantages in energy efficiency over ANNs, their performance on various benchmark tasks remains suboptimal. However, recent developments in the field of SNNs have shown notable progress. The introduction of surrogate gradient learning to circumvent the non-differentiability, introduced by the neuronal threshold mechanism \cite{neftci2019surrogate} allowed training of deep SNN \citep{zhou2023spikformer}, \citep{zhu2023spikegpt} and \cite{hu2024advancing}. 
Additional noteworthy advancements include optimizing the number of timesteps based on the visual complexity of inputs \cite{li2024seenn}, the introduction of a temporal-channel joint attention mechanism for SNNs \cite{zhu2024tcja}, and the co-optimization of neuronal parameters and synaptic delays alongside synaptic weights \cite{deckers2024co}.
These developments have advanced the performance of SNNs on classification tasks to a level that is increasingly comparable to that of ANNs.

\begin{figure}[h!]
    \centering
    \includegraphics[width=\linewidth]{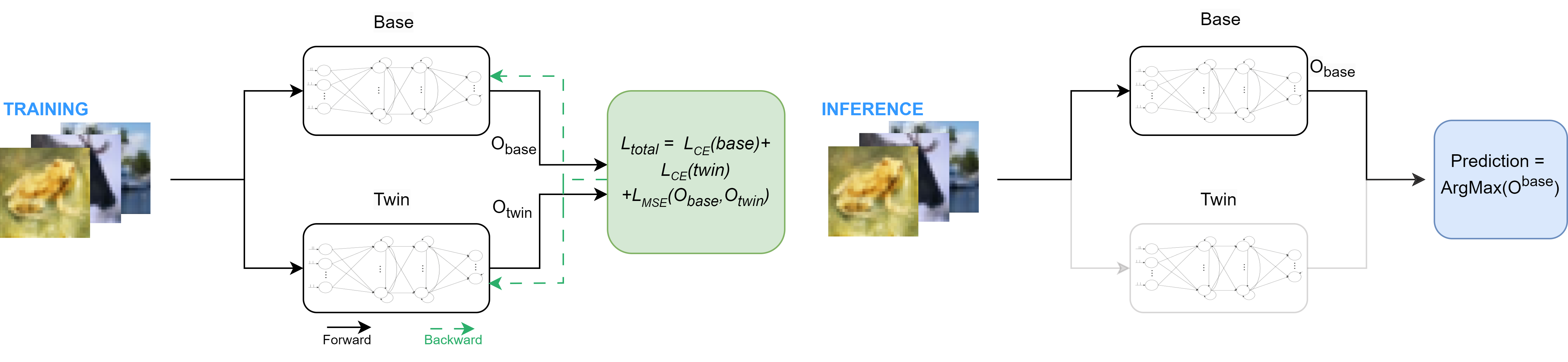}
    \caption{The overall workflow of the proposed novel algorithm is as follows: During the training phase, a twin network is instantiated and co-trained from scratch alongside the base SNN. The training loss comprises three components: the two individual cross-entropy losses of the two networks and the logit matching loss. In the inference phase, only the original base network is employed to generate predictions.}
    \label{augmented_net}
\end{figure}

In this study, we introduce Twin Network Augmentation (TNA) as an innovative training methodology for network regularization in SNNs. Specifically, during the training phase, a randomly initialized twin SNN with exactly the same network architecture is co-trained alongside the original base SNN. The training objective encompasses minimizing both their respective cross-entropy losses and the mean squared error between their output logits. This is the logit matching loss, which enforces convergence of the twin SNNs to similar network outputs, despite their distinct internal feature representations, thereby enhancing the expressivity of both models and providing regularization. During inference, only the original base SNN is utilized, resulting in a model that is both powerful and efficient.
The methodology of the proposed algorithm is illustrated in Fig. \ref{augmented_net}. This paper demonstrates that TNA 1) enhances the classification performance of SNNs across a range of (dynamic) vision datasets, achieving state-of-the-art results for the evaluated model architecture, and 2) Enhances classification performance when implemented in scenarios where the base SNN is compressed to ternary weight precision.

The contributions of this study are summarized as follows:

\begin{itemize} 
\item [1.] \quad We present TNA, a novel regularization technique for SNNs. TNA leverages multiple random initializations to provide diverse data views, thereby enhancing generalization performance and surpassing a comparable SNN optimized using knowledge distillation. 
\item [2.] \quad We demonstrate that TNA effectively guides SNNs when compressed to ternary weight precision, with the compressed SNN achieving superior performance compared to a full-precision SNN of the same architecture without TNA. 
\item [3.] \quad The proposed method is rigorously evaluated across several static datasets and one dynamic vision dataset, with TNA consistently improving performance for both full-precision and ternary weight networks. 
\end{itemize}

\section{Related work}

\textbf{Knowledge distillation}\\
Knowledge distillation (KD) \cite{hinton2015distilling}, \cite{gou2021knowledge} is a well-established technique for transferring knowledge from a large, fully trained teacher or ensemble model to a smaller student model, based on either the output logits or intermediate features \cite{romero2014fitnets}. KD has been employed to facilitate the transfer of knowledge from ANNs to SNNs \cite{xu2023constructing}, \cite{qiu2024self}, as well as from trained, full-precision networks to binary neural networks \cite{leroux2020training}, \cite{yang2021knowledge}. 

Unlike KD-based approaches, our method does not rely on a large pre-trained network. Instead, it addresses a different aspect: the performance variability across different initializations. We leverage this phenomenon by augmenting the spiking neural network with a co-trained twin network, minimizing the difference between their output logits. This approach mitigates implicit initialization bias \cite{ramasinghe2023much}, thereby enhancing classification performance.

\textbf{Siamese neural networks}\\
Siamese neural networks (SN) \cite{chicco2021siamese} represent a well-established approach in contrastive learning, characterized by a model architecture similar to the one proposed in this work. In SN, however, the twin networks share identical weights and undergo mirrored weight updates, with the objective of detecting similarities between their inputs. In contrast, the twin networks in our approach are trained with independent weight updates, and the optimization process is driven by a cross-entropy loss function, specifically designed to enhance the classification accuracy of one of the twin networks.

\textbf{Regularization methods}\\
Regularization methods refer to another class of works that address the prevalent issue of overfitting and can be categorized into two main types. The first category includes data augmentation techniques, such as Random Erasing \cite{zhong2020random}, Mixup \cite{zhang2018mixup}, and CutMix \cite{yun2019cutmix}, which transform input data to introduce noise into the dataset. The second category comprises Dropout-based methods \cite{srivastava2014dropout}, which introduce noise into the network architecture by randomly removing nodes, connections \cite{wan2013regularization}, or entire blocks in a structured manner based on their spatial correlations \cite{NEURIPS2018_7edcfb2d}. Additionally, network augmentation \cite{cai2022network} has been introduced as a method that enhances the performance of smaller models by embedding them within a larger model that shares weights and gradients.

In contrast to these regularization techniques, our proposed method does not seek to introduce noise into the dataset, the model, or augment it by embedding it within a larger model. Instead, our approach focuses on co-training the model alongside another model to obtain multiple 'views' of the same data. This strategy reduces reliance on a specific set of features, which are influenced by random model initialization, thereby enhancing the regularization of our SNN models after training.

\textbf{Binary/Ternary SNN}\\
In recent years, the combination of SNNs and binarization for low-power inference has been explored. \cite{lu2020exploring} introduced a method where a binary neural network is trained and subsequently converted into a binary SNN. Another study \cite{jang2021bisnn} proposed direct training of binary SNNs based on Bayesian principles. Additionally, \cite{kheradpisheh2022bs4nn} presented a technique for binarizing the weights of spiking networks to the range [-1, 1] during the forward pass in the training phase, incorporating time-to-first-spike encoding. More recently, adaptive local binary spiking neural networks \cite{pei2023albsnn} have been developed, which use local accuracy loss estimators to determine which layers should be binarized.

Ternary weight networks \cite{li2016ternary} were introduced as a memory- and computationally-efficient alternative to binary neural networks, offering enhanced expressive capability. These networks can be viewed as a sparser variant of binary networks, with some [-1, 1] values replaced by zero \cite{deng2022sparsity}. To address the balance between power and memory consumption, a sophisticated quantization policy for adaptive binary-ternary networks was proposed \cite{razani2021adaptive}. Recently, ternary weight spiking neural networks \cite{nguyen2022low} have also emerged as a promising candidate for low-power inference on neuromorphic chips.

In contrast to the binarization methods previously discussed, our approach begins with the training of a full-precision network, which is then co-trained using our novel model augmentation technique. After a predefined number of epochs, the network is compressed to ternary weights while continuing the training process. This final step enables the network to retain the benefits of model augmentation even after the compression to ternary weights.

\section{Preliminary}
\label{syn_learning}
For clarity, we will first introduce the neuron model, the leaky-integrate-and-fire (LIF) neuron as well surrogate gradients, which are typically used for training spiking neural networks (SNN). These concepts are commonly used for training spiking neural networks (SNNs) and are integral to this work.
\subsection{Neuron models}
In spiking neural networks (SNNs), the conventional activation function used in artificial neural networks (ANNs) is substituted with a temporally dependent function that emulates the spiking dynamics of biological neurons. A neuron model represents a mathematical formulation of the spatio-temporal integration performed by each neuron within the SNN. Formally, the leaky-integrate-and-fire (LIF) model is described as follows:

\begin{equation}
\label{LIF}
\begin{split}
& u[t] = \alpha (u[t-1] \cdot (1-s[t])) + I[t] \\
& s[t] = u[t] \geq \theta \\
\end{split}
\end{equation}
where the membrane potential u[t] decays over time with factor $\alpha$. When the membrane potential crosses the firing threshold $\theta$, a spike s[t] is generated at timestep $t$ and the membrane potential is reset to its resting potential, which is 0 in this case. The pre-synaptic weighted inputs are represented by $I[t]$. The spike trains s[t], which can thus only be either 0 or 1, are used to propagate the signal through the network. 

\subsection{Training SNN}

Spiking neural networks (SNNs) are commonly trained using backpropagation-through-time (BPTT) with surrogate gradients \citep{neftci2019surrogate}. In these approaches, the cross-entropy loss of the network is derived from the sum of the membrane potentials of the output neurons over the timesteps, with the network being unrolled in time. The loss with respect to class c for a batch size N is given by:
\begin{equation}
    \label{loss}
    \mathcal{L}_{c} = \frac{1}{N} \sum^{N}_{n=1}-log(\frac{e^{u_{out,c}}}{\sum_{j}e^{u_{out, j}}})
\end{equation}

Based on the chain rule in error-backpropagation, the weights $W$ are updated for neuron $i$ in the penultimate layer $l$ for a sequence of T timesteps as shown in Equation (\ref{weight_update0}). 

\begin{equation}
    \label{weight_update0}
    \frac{\delta \mathcal{L}_{c}}{\delta W^{l}} = \frac{1}{T}\sum^{T}_{t=1}\sum^{t}_{m=0}\frac{\delta \mathcal{L}_{c}[t]}{\delta u_{out}[m]} \frac{\delta u_{out}[m]}{\delta s_{l}[m]}\frac{\delta s_{l}[m]}{\delta u_{l}[m]} \frac{\delta u_{l}[m]}{\delta W_{l}} 
\end{equation}

Surrogate gradients are employed to address the issue of the non-differentiable nature of the Heaviside function, which arises from the spiking mechanism described in Equation \ref{LIF}. The surrogate gradient, $\frac{\delta s[t]}{\delta u[t]}$, is a differentiable approximation used during the backward pass of error backpropagation to estimate the actual gradient. For simplicity, this study utilizes the boxcar surrogate gradient function, as detailed in Equation (\ref{boxcar}). The output logits of the SNN models are obtained by summing the outputs across all timesteps.
\begin{equation}
\label{boxcar}
\frac{\delta s[t]}{\delta u[t]} =
\begin{cases}
0.5 & \text{if $|u[t] - \theta| \leq 0.5$}\\ 0 & \text{otherwise}
\end{cases}
\end{equation}

\section{Methodology}
In this paper, we postulate that one of the primary challenges in training highly accurate spiking neural networks—namely, their suboptimal generalization performance—can be addressed through a novel approach called twin network augmentation. This method involves co-training a twin SNN from scratch alongside the original network. Additionally, we extend this novel training technique to enhance sparse binarization with ternary weights for SNNs. We first outline the novel training method and subsequently describe its application to compress SNNs to ternary weight precision.

\subsection{Twin network augmentation in spiking neural networks}
\label{model_augment}
We define a base spiking neural network $N_{base}$ with initial weights $W_{base}^{init}$. Using the cross-entropy loss function $\mathcal{L}$ as specified in Equation \ref{loss}, and applying the weight update rule outlined in Equation \ref{weight_update0}, the weights are iteratively updated according to the equation $W_{base}^{n+1} = W_{base}^{n} - \eta\frac{\delta \mathcal{L}c}{\delta W{base}}$, where $\eta$ represents the learning rate. Under the assumption of standard stochastic gradient descent, the initialization of the weights defines the starting point for the optimization process, which plays a critical role in ensuring convergence and the subsequent performance of the network.

This paper aims to enhance the performance of spiking neural networks (SNNs) by co-training a twin network, denoted as $N^{twin}$, alongside the original SNN, $N_{base}$. The optimization involves minimizing the mean squared error (MSE) between the output logits of both networks, referred to as the logit matching loss. This approach integrates the initialization of both networks and aggregates the distinct 'views' of the input data, which are utilized during the backward pass for both networks. The formal definition of the proposed loss is provided in Equation \ref{augmented_loss}.

\begin{equation}
    \label{augmented_loss}
    \mathcal{L}_{total}= \underbrace{\mathcal{L}_{CE}(W^{base})}_\text{Base loss}+ \underbrace{\mathcal{L}_{CE}(W^{twin})}_\text{Twin loss} +
     \alpha \cdot \underbrace{\mathcal{L}_{MSE}(O^{base}, O^{twin})}_\text{Logit matching loss}
\end{equation}

where $\mathcal{L}_{CE}$ denotes the cross-entropy loss for the base and twin network. The logit matching loss is the mean square error between the logits of the base network and the twin network, summed over all available timesteps. Lastly, we define $\alpha$ as a hyperparameter, used to balance the training loss of the individual SNN and the logit matching loss. Fine-tuning of $\alpha$ is necessary for every dataset and will be elaborated upon in Section \ref{alpha}. Note that both networks are trained from scratch and that the twin network is only used in the training phase. In inference, only the original base network is used. Figure \ref{augmented_net} shows the entire novel procedure in both the training and inference phase.

\subsection{Ternary spiking neural networks}
\label{ternary_SNN}

Ternary weight networks are a class of neural networks in which the weights are restricted to the set {-1, 0, 1}, reducing the weight resolution from the standard 32 bits (full precision) to 2 bits. This weight discretization helps address the performance degradation commonly associated with binary neural networks when compared to full-precision models. Additionally, the energy-efficient characteristics of spiking neural networks (SNNs) can be further leveraged by compressing the model to ternary weights. The compression of the weight matrix $W^{l}$ in layer $l$ using a symmetric threshold $\Delta$ is defined in Equation \ref{ternary}.

\begin{equation}
\label{ternary}
W^{l}_{t} =
\begin{cases}
-1 & \text{$W^{l}  <- \Delta $}\\
0 & \text{$|W^{l}|  \leq \Delta $}\\
+1 & \text{$W^{l}  > \Delta $}\\
\end{cases}
\end{equation}

Determining the appropriate value for $\Delta$ is a non-trivial task. In our experiments, we empirically set $\Delta = 0.1$. Following the common practice in binarizing spiking neural networks \cite{deng2021comprehensive}, we do not reduce the precision of the first and last layers to ternary weights. The first layer functions as a spike encoding layer, while the final layer requires full precision for sufficient expressivity. Reducing the precision of these layers significantly degrades model performance. Unlike previous approaches, where networks are quantized to binary or ternary precision from the outset, we delay the compression to ternary weights until after a fixed number of training epochs. Initially, during the exploration phase of learning, we maintain the model in full precision. Once the network training has converged, we compress the model to ternary weights, allowing fine-tuning at this reduced precision. Notably, even in spiking neural networks, where activations are already binarized, we observed performance improvements in the training phase, even after model compression to ternary weights.

Additionally, the twin network augmentation (TNA) training method, introduced in Section \ref{model_augment}, can be applied to achieve a more regularized compressed model. This approach involves training both a base and a twin spiking neural network (SNN). After the initial training phase, the base SNN is compressed to ternary weight precision, while the twin network retains full-precision weights. The co-training of both networks is then resumed, with the base and twin models optimized together until a predefined stopping criterion is met.

\section{Experiments}
In this section, we provide a detailed description of the experiments conducted in this study and the corresponding results. First, we outline the datasets used and the specific setup employed for training the spiking neural networks (SNNs). Next, we analyze the performance of the proposed methods, followed by a comprehensive comparison of our results with state-of-the-art SNN performance. 

For consistency with prior work, we selected the widely-adopted CIFARNet architecture \cite{wu2019direct} for all experiments. We incorporated Dropout \cite{srivastava2014dropout} into the fully connected layers and trained all SNNs for 5 timesteps, with the output computed as the sum of the membrane potentials across all timesteps.
\subsection{Setup}
\subsubsection{Datasets}

The experiments were conducted on widely-used non-spiking datasets, including CIFAR-10, CIFAR-100, and Fashion-MNIST \cite{xiao2017/online}. Additionally, the CIFAR-10-DVS dataset \cite{li2017CIFAR10} was included as a popular neuromorphic dataset, which uses event-based inputs and is compatible with neuromorphic hardware.

Standard preprocessing and data augmentation techniques were applied to the non-spiking datasets, consisting of the following steps: (1) padding the original image by 4 pixels, (2) applying RandomCrop, (3) horizontally flipping the image with a 50\% probability, and (4) normalizing the image using the mean (0.4914, 0.4822, 0.4465) and standard deviation (0.2470, 0.2435, 0.2616) across the RGB channels. During evaluation, only normalization was applied.
For the CIFAR-10-DVS dataset, the augmentation pipeline included: (1) RandomRotation with a rotation range of 30 degrees, and (2) RandomAffine with no rotation but a shear range of (-30, 30) degrees.

\subsection{Training details}
For all experiments, we employed the Adam optimizer \cite{kingma2014adam} with an initial learning rate of 0.01 and a batch size of 256, running on a Tesla V100-SXM2-32GB GPU. Each network was trained for 250 epochs, and if applicable, compression to binary or ternary weights commenced after 150 epochs. A learning rate scheduler was used, reducing the learning rate at each step by a factor of $\gamma=0.928$.

We introduced Dropout with a probability of 0.2 for the non-spiking datasets and 0.5 for the CIFAR-10-DVS dataset, applied to the fully connected layers. All networks were initialized using the default Kaiming initialization \cite{he2015delving} from PyTorch \cite{paszke2019pytorch}. For each dataset, the hyperparameter $\alpha$ was fine-tuned by evaluating values in the range $[1.e^{-2}, 1.e^{-3}, \dots, 1.e^{-6}]$. The impact of $\alpha$ is discussed in Section \ref{alpha}. Since the goal of this study is to demonstrate the effectiveness of the proposed training method, rather than achieving state-of-the-art performance, other hyperparameters were not fine-tuned.

\subsection{Results}

\subsubsection{Network augmentation effectively regularizes SNN}

The objective of our first experiment is to validate the hypothesis that co-training a twin spiking neural network (SNN) enhances the performance of the base SNN. To this end, we compare the training outcomes of a standalone baseline CIFARNet SNN with those obtained using our proposed twin network augmentation (TNA) method. The results across the four datasets are summarized in Table \ref{tab:augment_eval}. The findings demonstrate that the application of TNA consistently improves classification performance across all datasets. Notably, the improvements are more pronounced on more challenging datasets, such as CIFAR-100 and CIFAR-10-DVS.
\begin{table}[h!]
 \caption{Network augmentation consistently improves the CIFARNet accuracy across all evaluated datasets. The difference, denoted by $\Delta$, is more prominent in more challenging datasets, CIFAR-100 and CIFAR-10-DVS.}
\begin{center}
% \resizebox{\textwidth}{!}{%}
\begin{tabular}{c |c |c |c |c}
 Dataset & CIFAR-10 & CIFAR-100 & Fashion-MNIST & CIFAR-10-DVS  \\ [0.5pt]
\hline 
$\alpha$ & $1.e^{-3}$ & $1.e^{-3}$& $1.e^{-4}$ & $1.e^{-4}$\\[0.5pt]
\hline
Baseline  & 93.57\% & 72.6\%  & 94.90\% & 71.7\%\\
Twin augment.  & 94.39\% & 75.0\%  & 95.31\% & 73.4\% \\
\hline
$\Delta$ acc.  & \textbf{+0.82}\% & \textbf{+2.4\%}  & \textbf{+0.41\%} & \textbf{+1.7\%}  \\

\hline
\end{tabular}
\end{center}
\label{tab:augment_eval}
\end{table}

We also examined the classification performance differences between the base SNN and the twin network. Despite initializing the two networks differently, we observed no statistically significant difference in their performance. This finding supports the decision to select either of the two models for further use. The logit matching loss effectively ensures that the output logits of both models are closely aligned.

In a further experiment to assess the effectiveness of the proposed twin network augmentation, we extended the approach to include three networks instead of just the base and twin SNNs. The loss function was modified to incorporate an additional cross-entropy loss and logit matching loss relative to the target SNN. However, this configuration did not yield any statistically significant performance improvements compared to the dual SNN setup.
\subsubsection{The role of the balancing parameter alpha}
\label{alpha}

\begin{figure} [H]
\centering
\begin{tabular}{c c c}
\includegraphics[width=0.3\textwidth]{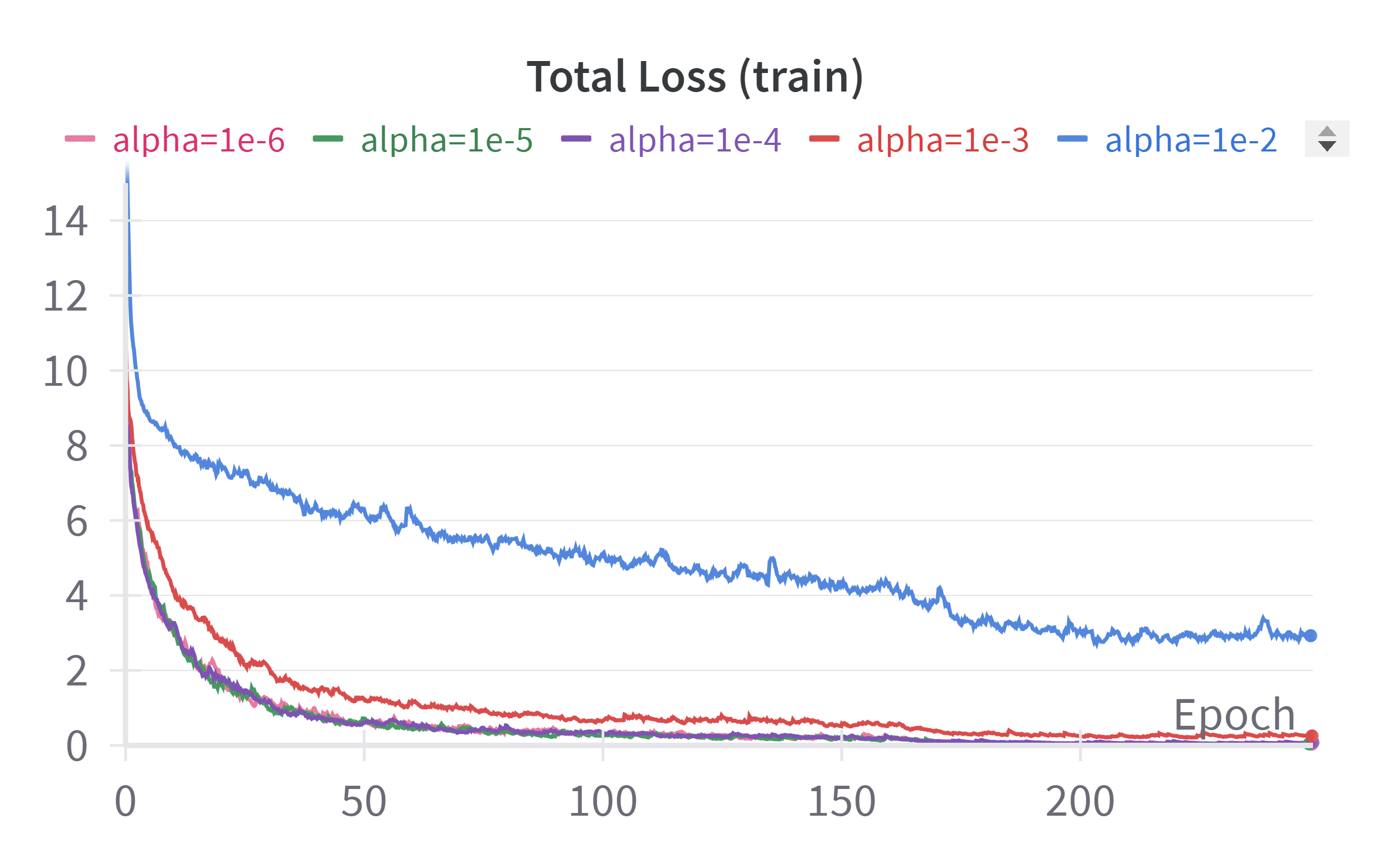} &
\includegraphics[width=0.3\textwidth]{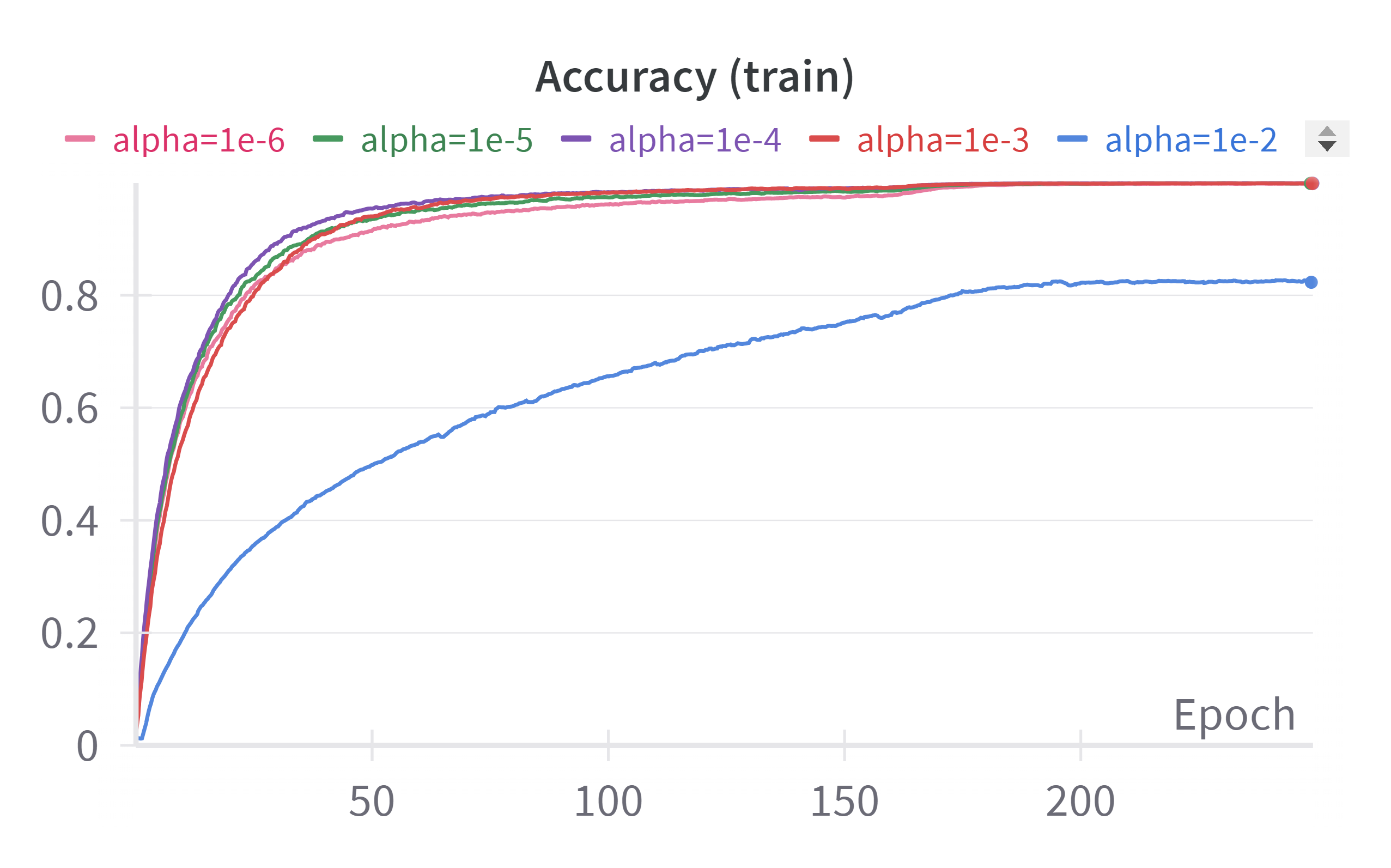} &
\includegraphics[width=0.3\textwidth]{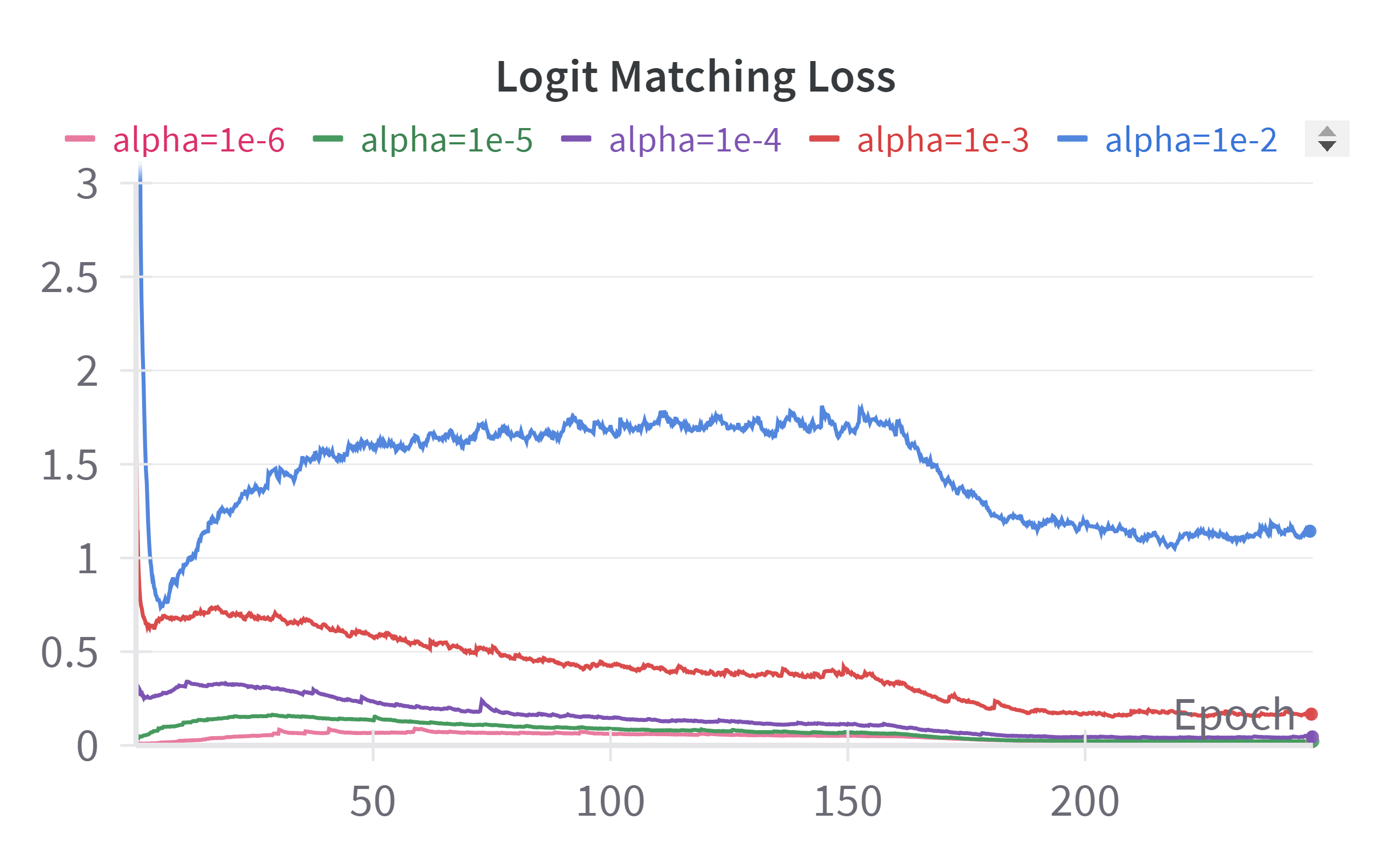} \\
\textbf{(a)}  & \textbf{(b)} & \textbf{(c)} \\
\includegraphics[width=0.3\textwidth]{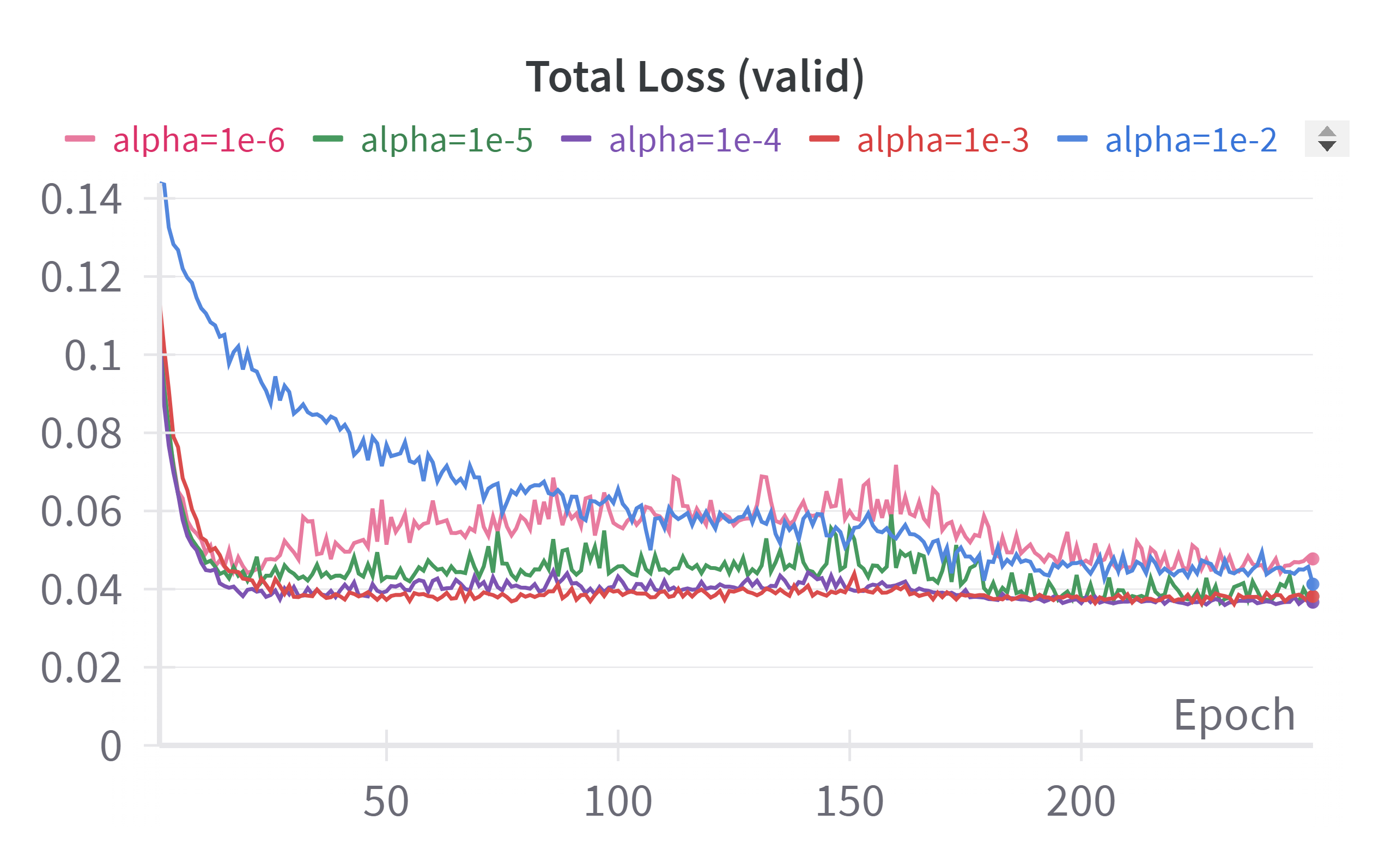} &
\includegraphics[width=0.3\textwidth]{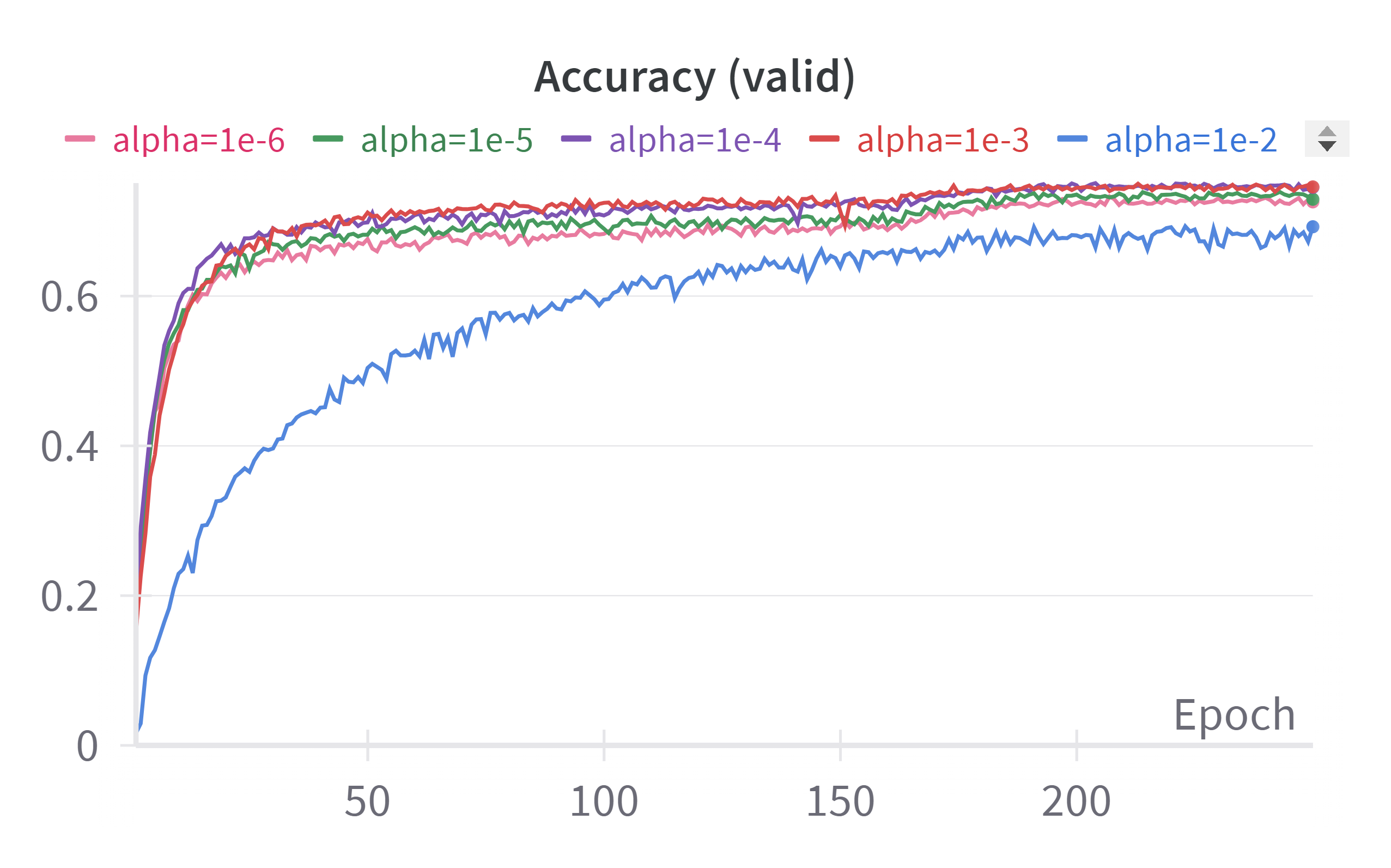} \\
\textbf{(d)} & \textbf{(e)}  \\
\end{tabular}
\caption{The significance of selecting an appropriate balancing parameter $\alpha$. We show the loss on the CIFAR-100 training \textbf{(a)} and validation set \textbf{(d)} as well as the classification accuracy on the training set \textbf{(b)} and validation set \textbf{(e)}. In \textbf{(c)} we show the magnitude of the matching loss.}
\label{fig:alphafig}
\end{figure}

As defined in Equation \ref{augmented_loss}, the hyperparameter $\alpha$ regulates the trade-off between the cross-entropy losses for training the base and twin SNNs and the logit matching loss. The optimal value of $\alpha$ is influenced by both the batch size and the dimensions of the logits matrix. When $\alpha$ is set too high, the logit matching loss may dominate, hindering the training of both individual SNNs. Conversely, if $\alpha$ is set too low, the advantages of co-training the twin SNN are diminished. The next experiment investigates the impact of $\alpha$ on the training process. Figure \ref{fig:alphafig} illustrates the cross-entropy loss and logit matching loss on both the CIFAR-100 training and validation sets, along with the classification accuracy, for various values of $\alpha$. Similar behavior was observed for the other datasets.

We identify two primary observations from our analysis. First, as depicted in Fig. \ref{fig:alphafig}, a high value of $\alpha$ (denoted in blue) impedes the learning process. Specifically, the logit matching loss increases continuously throughout training, indicating that it overly influences the total loss and hampers the effective training of the SNNs. Conversely, a low value of $\alpha$ fails to reveal the benefits of co-training the twin SNN. Second, an examination of the logit matching loss (see Fig. \ref{fig:alphafig} C) across all $\alpha$ values shows an initial decrease, which corresponds to the learning of basic and fundamental features. This is followed by a slight increase in the matching loss, which subsequently decreases during the fine-tuning phase of the SNN models.

\subsubsection{Comparison to Knowledge Distillation}

\begin{figure} [H]
\centering
\begin{tabular}{c c}
\includegraphics[width=0.33\textwidth]{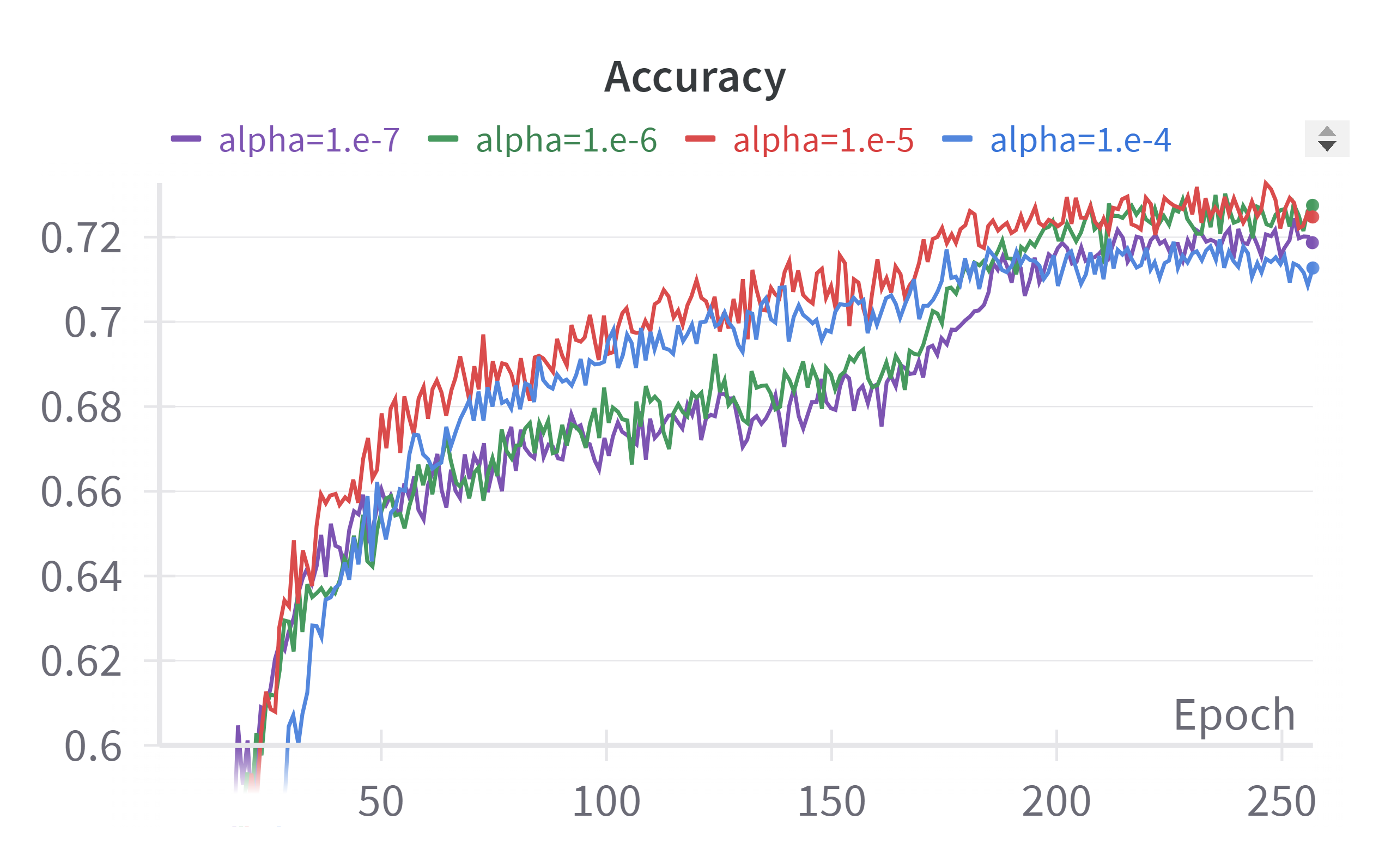} &
\includegraphics[width=0.33\textwidth]{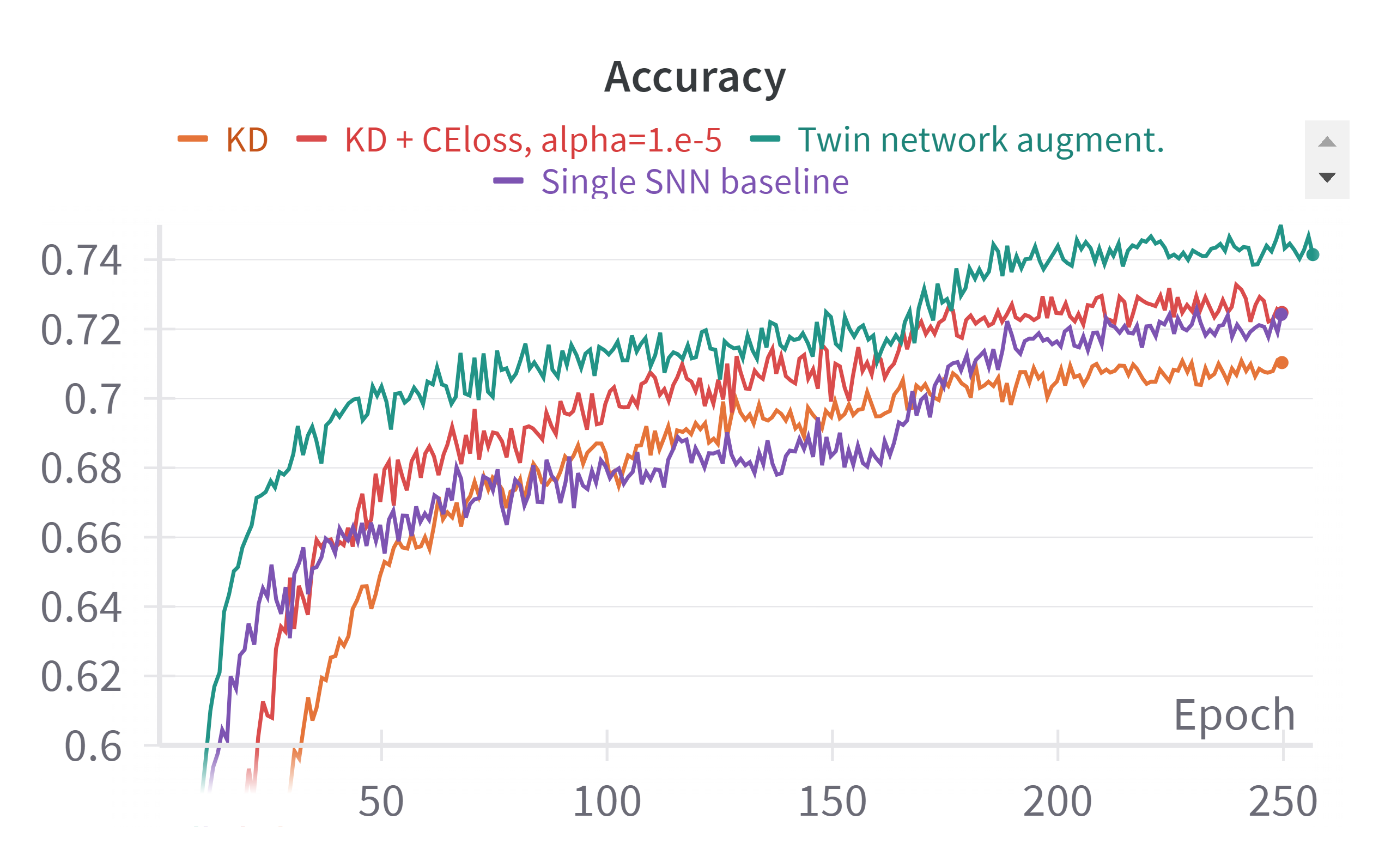} \\
\textbf{(a)} Influence of $\alpha$ on the KD + CEloss model  & \textbf{(b)} Comparison with the other methods. \\[6pt]
\end{tabular}
\caption{\textbf{(a)} Hyperparameter $\alpha$ is used to balance between the CEloss and the KD logit matching loss. We found an $\alpha$ of $1.e^{-5}$ performs best in this particular scenario. 
\textbf{(b)} Comparison of the KD-based methods with our baseline SNN and the new twin network augmentation SNN on the CIFAR-100 dataset.}
\label{fig:kd_comp}
\end{figure}

The method most analogous to our novel twin network augmentation approach is Knowledge Distillation (KD). In this section, we compare the performance of our proposed method against: 1) the single baseline SNN, 2) KD + CEloss: an SNN trained with a pre-trained twin network that is not co-trained with the base SNN. This configuration results in a loss function similar to Equation \ref{augmented_loss} but without the twin loss term, as the twin SNN is not trained concurrently, and 3) KD: a pure knowledge distillation approach where only the logit matching loss is used to train the base SNN. The results of these comparisons on the CIFAR-100 dataset are presented in Fig. \ref{fig:kd_comp}.

To compare our augmentation method with similar approaches, we first conducted experiments to identify the optimal $\alpha$ for the KD + CELoss configuration. Figure \ref{fig:kd_comp} (a) displays the convergence curves for different $\alpha$ values, with $\alpha = 1.e^{-5}$ proving to be the most effective. Figure \ref{fig:kd_comp} (b) demonstrates that our twin network augmentation (with $\alpha = 1.e^{-4}$) significantly surpasses traditional knowledge distillation (KD) with a twin SNN model by 3.93\%, and also outperforms the KD-based teacher model with additional cross-entropy loss for the base SNN by 1.70\%. Furthermore, the KD model with cross-entropy loss shows a 0.73\% improvement over the single SNN baseline, indicating that fixed additional supervision enhances the training process. This suggests the potential benefits of exploring co-training with larger 'sibling' models rather than relying solely on an identical twin model, as is typically done in knowledge distillation. Lastly, the baseline model outperforms the KD model, as the latter did not have access to the ground truth labels.

\subsubsection{Evaluation of the proposed SNN quantization method}

\textbf{T-SNN vs binary SNN}

We first assess the performance of the proposed spiking neural network (SNN) with ternary weights (T-SNN) in comparison to full-precision (FP) SNNs and binary SNNs (B-SNN) with weights in {-1, 1}. The results, as presented in Table \ref{tab:evalTNA_TSNN}, indicate that T-SNN significantly outperforms B-SNN.
We observe two primary trends when reducing weight precision to ternary values. First, the accuracy loss relative to full-precision weights is minimal, with a notable exception for the CIFAR-100 dataset, where a substantial accuracy drop is observed. This drop may be attributed to the need for more fine-grained features in this more complex dataset.
Second, the T-SNN achieves classification accuracy comparable to that of the full-precision model on the CIFAR-10 and Fashion-MNIST datasets, demonstrating that ternary weights effectively balance classification performance with computational efficiency.

\begin{table}[h!]
 \caption{Top:The proposed compression methods compared to full-precision (FP). The ternary weight network (T-SNN) consistently outperforms it binary counterpart (B-SNN) and approaches the FP SNN. Bottom: TNA clearly improves oerformance of the resulting compressed T-SNN, which achieves similar performance as the FP-SNN model on CIFAR10/100 and outperforms it on Fashion-MNIST and CIFAR-10-DVS.}
 
\begin{center}
% \resizebox{\textwidth}{!}{%}
\begin{tabular}{c |c |c |c |c}
 Dataset & CIFAR-10 & CIFAR-100 & Fashion-MNIST & CIFAR-10-DVS  \\ [0.5pt]
\hline 
$\alpha$ & $1.e^{-3}$ & $1.e^{-3}$& $1.e^{-4}$ & $1.e^{-4}$\\[0.5pt]
\hline
FP-SNN  & 93.57\% & 72.6\%  & 94.90\% & 71.7\%\\
B-SNN & 92.31\% & 69.85\%  & 94.80\% & 70.6\%\\
T-SNN & 92.95\% & 70.52\% & 94.95\% & 71.3\%\\ [0.5pt]
\hline
TNA FP-SNN  & 94.39\% & 75.0\%  & 95.31\% & 73.4\% \\
TNA T-SNN  & 93.23\%  & 72.03\% & 95.24\% & 72.2\% \\
\hline
\end{tabular}
\end{center}
\label{tab:evalTNA_TSNN}
\end{table}

\textbf{Exploiting twin network augmentation for ternary weight precision SNN}

Twin network augmentation can also be applied to supervise spiking neural networks (SNNs) compressed to ternary weights (T-SNN), as detailed in Section \ref{ternary_SNN}. In this section, we compare the performance of T-SNN with twin network augmented T-SNN (TNA T-SNN) against their full-precision (FP) counterparts. The results are summarized in Table \ref{tab:evalTNA_TSNN}.

Our findings demonstrate that employing TNA during the compression of SNNs to ternary weights, while retaining the twin SNN in full precision, substantially enhances model performance across all evaluated datasets. As observed in comparisons with full-precision models, the most significant improvements occur with the CIFAR-100 dataset. Additionally, it is noteworthy that the low-precision TNA T-SNN models achieve performance comparable to that of the single full-precision SNN models across all datasets and even outperform them on Fashion-MNIST and CIFAR-10-DVS.

\subsubsection{Comparison to the state-of-the-art models}

Table \ref{TNA_full_results} presents the performance of the models proposed in this study compared to a selection of state-of-the-art spiking neural network (SNN) models. For a fair comparison, we have also included results from convolutional neural network (CNN) models of comparable or greater size. Results for full-precision models are shown above the dotted line, while those for ternary-weight SNNs (T-SNN) are displayed below. Given the limited number of available ternary-weight SNN models, we have also included results from recent binary SNN studies where available, denoted by $^{*}$. Some references include results from non-standard CNN architectures, which are labeled as CNN1, CNN2, and CNN3 in Table \ref{TNA_full_results}.

\begin{table}[h!]
 \caption{Comparison of the proposed twin augmentation (TNA) SNN and the TNA T-SNN, reduced to ternary weight precision to the state of the art in SNN and binary/ternary SNN respectively. Our results are highlighted in bold.}
\begin{center}
\begin{tabular}{llllr}
\hline
   Dataset &  Method & Architecture &  timestep & Accuracy  \\  
 \hline\hline
 & IM-Loss \cite{guo2022loss} & CIFARNet  & 4 & 92.20\% \\  
 CIFAR-10  & DSR \cite{fang2021incorporating} & ResNet-18  & 20 & 95.40\% \\ 
 & IM-Loss \cite{guo2022loss} & ResNet-19  & 6 & 95.49\% \\  
 & \textbf{TNA SNN} (ours) &  CIFARNet & 5 & \textbf{94.39}\% \\

\hdashline
& GAND-Nets \cite{wu2019direct} & ResNet14-64 & 12 & 87.42\% \\
& TW-SNN \cite{nguyen2022low} & VGG-16 & 250 & 89.71\% \\
& Asymmetric ResNeXt \cite{wu2023hardware} & ResNext & 16 & 90.00\% \\
& ReRam \cite{lin2020scalable} & SResNet & 700 & 92.1\% \\

 & \textbf{TNA T-SNN} (ours) &  CIFARNet& 5 & \textbf{93.23}\%\\ [0.5ex]  
 \hline 
& IM-Loss \cite{guo2022loss} & VGG-16 & 5 & 70.18\%\\
 CIFAR-100 & DSR \cite{fang2021incorporating} & ResNet-18  & 20 & 78.50\% \\  
 & \textbf{TNA SNN} (ours) & CIFARNet & 5 & \textbf{75.0}\%\\

\hdashline
 & B-SNN$^{*}$ \cite{lu2020exploring} & VGG-16 & $>$200 & 63.07\% \\
 & GAND-Nets \cite{wu2019direct} & ResNet18-64 & 12 & 63.42\% \\
 & ALBSNN$^{*}$ \cite{pei2023albsnn} & CIFARNet & 1 & 69.55\%\\
  & NUTS-BSNN$^{*}$ \cite{dinh2023nuts} & CNN1 & 14 & 70.31\% \\
 & \textbf{TNA T-SNN} (ours) & CIFARNet& 5 & \textbf{72.03}\%\\ [0.5ex]  
\hline
 & PLIF \cite{fang2021incorporating} & CNN2 & 8 & 94.65\% \\
  Fashion-MNIST & TCJA-SNN \cite{zhu2024tcja} & CNN2 & 8 & 94.8\% \\
 & \textbf{TNA SNN} (ours) & CIFARNet & 5 & 95.31\% \\

\hdashline
 & BS4NN$^{*}$ \cite{kheradpisheh2022bs4nn} & MLP & 256 & 87.5\% \\
 & NUTS-BSNN$^{*}$ \cite{dinh2023nuts} & CNN3 & 14 & 93.25\% \\
 & \textbf{TNA T-SNN} (ours) & CIFARNet& 5 & \textbf{95.23}\%\\ [0.5ex]  
 
 \hline
 & RecDis-SNN \cite{guo2022recdis} & CIFARNet & - & 67.30\% \\
CIFAR-10-DVS& IM-Loss \cite{guo2022loss} & ResNet-19 & 10 & 72.60\%\\
 & DSR \cite{fang2021incorporating} & VGGSNN  & 20 & 77.27\% \\  
  & TET \cite{deng2022temporal} & VGGSNN & 10 & 77.33\% \\
  & \textbf{TNA SNN} (ours) & CIFARNet & 5 & \textbf{73.4}\% \\

\hdashline
 & ALBSNN$^{*}$ \cite{pei2023albsnn} & CIFARNet & 1 & 68.98\%\\
 & \textbf{TNA T-SNN} (ours) & CIFARNet & 5 & \textbf{72.2}\%\\ [0.5ex]  

 \hline
 *Binary SNN
\end{tabular}
\end{center}
\label{TNA_full_results}
\end{table}

We begin by comparing our full-precision (FP) results with those of other works. Our analysis reveals that, on the CIFAR-10, CIFAR-100, and CIFAR-10-DVS datasets, the accuracy of the twin network augmented SNN (TNA SNN) is close to the state-of-the-art, despite the CIFARNet architecture (128C3-256C3-AP2-512C3-AP2-1023C3-512C3 - 1024FC-512FC-Out) being significantly simpler and smaller than the models used in other studies (VGGs and ResNets). On the Fashion-MNIST dataset, the TNA SNN achieves state-of-the-art performance. When compared with other studies using the same model architecture, the TNA SNN demonstrates substantial improvements: +2.19\% on CIFAR-10 and +6.1\% on CIFAR-10-DVS. Notably, the TNA SNN based on the CIFARNet architecture surpasses a VGG-16-based model on the CIFAR-100 dataset, despite the VGG-16 architecture being shown to outperform CIFARNet in \cite{guo2022loss}.

Subsequently, we compare the TNA ternary-weight SNN (T-SNN) against state-of-the-art binary and ternary SNNs. Our proposed model surpasses all existing benchmarks across all datasets in the binary and ternary SNN categories, demonstrating that our twin network augmentation combined with ternary weight resolution is a highly effective approach for SNN compression.

\section{Conclusion}

In this paper, we introduced a novel training strategy for spiking neural networks (SNNs), termed twin network augmentation (TNA). This approach involves co-training a duplicate of the original SNN architecture, initialized with different weights, alongside the base SNN. During inference, only the base SNN is retained. We demonstrated that TNA offers two primary benefits: (1) it enhances network regularization by incorporating multiple weight initializations, leading to improved classification accuracy and reduced initialization bias, and (2) it facilitates better quantization of SNNs to ternary-weight precision by co-training the quantized SNN with a full-precision counterpart. Moreover, we showed that TNA outperforms a comparable model trained using knowledge distillation.

Our results highlighted that TNA SNN achieve state-of-the-art performance across benchmark datasets, particularly when compared to other SNNs of similar size, and even surpasses larger networks in some cases. Additionally, the ternary-weight [-1, 0, 1] SNN (T-SNN) trained with TNA demonstrated superior performance over all available binary and ternary-weight SNNs.

This research primarily focuses on the CIFARNet SNN model and (dynamic) vision datasets. Future research could extend the twin network augmentation method to a broader range of datasets and model architectures. In this study, we employed an exact replica of the SNN for co-training, but an interesting direction for future exploration would be to assess the impact of co-training larger or smaller models. Similar to advancements in knowledge distillation, it would be valuable to investigate how incorporating more internal network features into the logit matching loss could further improve performance. While adding a third SNN to the TNA process did not yield additional benefits, this could change when co-training with larger models or different architectures, drawing parallels to the success of mixture-of-experts models \cite{NEURIPS2022_2f00ecd7}. Finally, extending twin network augmentation to non-spiking artificial neural networks (ANNs) represents another promising avenue for future work.

\section*{Conflict of Interest Statement}

The authors declare that the research was conducted in the absence of any commercial or financial relationships that could be construed as a potential conflict of interest.

\section*{Author Contributions}

LD and BV contributed to development and design of the experiments. LD performed the experiments and the analysis. All authors contributed to manuscript writing, revision, read, and approved the submitted version.

\section*{Funding}
This work was supported by a SB grant (1S87022N) from the Research Foundation — Flanders (FWO).
 
\section*{Data Availability Statement}
All datasets presented in the study are publicly available. The datasets that were used to evaluate the findings of this study are openly available: Fashion-MNIST at \url{https://doi.org/10.48550/arXiv.1708.07747},
CIFAR-10 at \url{http://www.cs.utoronto.ca/~kriz/cifar.html}, CIFAR-100 at \url{http://www.cs.utoronto.ca/~kriz/cifar.html} and CIFAR-10-DVS
at \url{https://doi.org/10.3389/fnins.2017.00309}.

\section*{References}
\bibliographystyle{plainnat}

\bibliography{TNA_v1}

\begin{thebibliography}{53}
\providecommand{\natexlab}[1]{#1}
\providecommand{\url}[1]{\texttt{#1}}
\expandafter\ifx\csname urlstyle\endcsname\relax
  \providecommand{\doi}[1]{doi: #1}\else
  \providecommand{\doi}{doi: \begingroup \urlstyle{rm}\Url}\fi

\bibitem[Bouvier et~al.(2019)Bouvier, Valentian, Mesquida, Rummens, Reyboz, Vianello, and Beigne]{bouvier2019spiking}
Maxence Bouvier, Alexandre Valentian, Thomas Mesquida, Francois Rummens, Marina Reyboz, Elisa Vianello, and Edith Beigne.
\newblock Spiking neural networks hardware implementations and challenges: A survey.
\newblock \emph{ACM Journal on Emerging Technologies in Computing Systems (JETC)}, 15\penalty0 (2):\penalty0 1--35, 2019.

\bibitem[Cai et~al.(2022)Cai, Gan, Lin, and Han]{cai2022network}
Han Cai, Chuang Gan, Ji~Lin, and Song Han.
\newblock Network augmentation for tiny deep learning.
\newblock In \emph{International Conference on Learning Representations}, 2022.
\newblock URL \url{https://openreview.net/forum?id=TYw3-OlrRm-}.

\bibitem[Chicco(2021)]{chicco2021siamese}
Davide Chicco.
\newblock Siamese neural networks: An overview.
\newblock \emph{Artificial neural networks}, pages 73--94, 2021.

\bibitem[Courbariaux et~al.(2015)Courbariaux, Bengio, and David]{courbariaux2015binaryconnect}
Matthieu Courbariaux, Yoshua Bengio, and Jean-Pierre David.
\newblock Binaryconnect: Training deep neural networks with binary weights during propagations.
\newblock \emph{Advances in neural information processing systems}, 28, 2015.

\bibitem[Deckers et~al.(2024)Deckers, Van~Damme, Van~Leekwijck, Tsang, and Latr{\'e}]{deckers2024co}
Lucas Deckers, Laurens Van~Damme, Werner Van~Leekwijck, Ing~Jyh Tsang, and Steven Latr{\'e}.
\newblock Co-learning synaptic delays, weights and adaptation in spiking neural networks.
\newblock \emph{Frontiers in Neuroscience}, 18:\penalty0 1360300, 2024.

\bibitem[Deng et~al.(2021)Deng, Wu, Hu, Liang, Li, Hu, Ding, Li, and Xie]{deng2021comprehensive}
Lei Deng, Yujie Wu, Yifan Hu, Ling Liang, Guoqi Li, Xing Hu, Yufei Ding, Peng Li, and Yuan Xie.
\newblock Comprehensive snn compression using admm optimization and activity regularization.
\newblock \emph{IEEE transactions on neural networks and learning systems}, 34\penalty0 (6):\penalty0 2791--2805, 2021.

\bibitem[Deng et~al.(2022)Deng, Li, Zhang, and Gu]{deng2022temporal}
Shikuang Deng, Yuhang Li, Shanghang Zhang, and Shi Gu.
\newblock Temporal efficient training of spiking neural network via gradient re-weighting.
\newblock In \emph{International Conference on Learning Representations}, 2022.
\newblock URL \url{https://openreview.net/forum?id=_XNtisL32jv}.

\bibitem[Deng and Zhang(2022)]{deng2022sparsity}
Xiang Deng and Zhongfei Zhang.
\newblock Sparsity-control ternary weight networks.
\newblock \emph{Neural Networks}, 145:\penalty0 221--232, 2022.

\bibitem[Dinh et~al.(2023)Dinh, Bui, Nguyen, John, Lin, and Trinh]{dinh2023nuts}
Van-Ngoc Dinh, Ngoc-My Bui, Van-Tinh Nguyen, Deepu John, Long-Yang Lin, and Quang-Kien Trinh.
\newblock Nuts-bsnn: A non-uniform time-step binarized spiking neural network with energy-efficient in-memory computing macro.
\newblock \emph{Neurocomputing}, 560:\penalty0 126838, 2023.

\bibitem[Fang et~al.(2021)Fang, Yu, Chen, Masquelier, Huang, and Tian]{fang2021incorporating}
Wei Fang, Zhaofei Yu, Yanqi Chen, Timoth{\'e}e Masquelier, Tiejun Huang, and Yonghong Tian.
\newblock Incorporating learnable membrane time constant to enhance learning of spiking neural networks.
\newblock In \emph{Proceedings of the IEEE/CVF international conference on computer vision}, pages 2661--2671, 2021.

\bibitem[Ghiasi et~al.(2018)Ghiasi, Lin, and Le]{NEURIPS2018_7edcfb2d}
Golnaz Ghiasi, Tsung-Yi Lin, and Quoc~V Le.
\newblock Dropblock: A regularization method for convolutional networks.
\newblock In S.~Bengio, H.~Wallach, H.~Larochelle, K.~Grauman, N.~Cesa-Bianchi, and R.~Garnett, editors, \emph{Advances in Neural Information Processing Systems}, volume~31. Curran Associates, Inc., 2018.
\newblock URL \url{https://proceedings.neurips.cc/paper_files/paper/2018/file/7edcfb2d8f6a659ef4cd1e6c9b6d7079-Paper.pdf}.

\bibitem[Gou et~al.(2021)Gou, Yu, Maybank, and Tao]{gou2021knowledge}
Jianping Gou, Baosheng Yu, Stephen~J Maybank, and Dacheng Tao.
\newblock Knowledge distillation: A survey.
\newblock \emph{International Journal of Computer Vision}, 129\penalty0 (6):\penalty0 1789--1819, 2021.

\bibitem[Guo et~al.(2022{\natexlab{a}})Guo, Chen, Zhang, Liu, Wang, Huang, and Ma]{guo2022loss}
Yufei Guo, Yuanpei Chen, Liwen Zhang, Xiaode Liu, Yinglei Wang, Xuhui Huang, and Zhe Ma.
\newblock Im-loss: information maximization loss for spiking neural networks.
\newblock \emph{Advances in Neural Information Processing Systems}, 35:\penalty0 156--166, 2022{\natexlab{a}}.

\bibitem[Guo et~al.(2022{\natexlab{b}})Guo, Tong, Chen, Zhang, Liu, Ma, and Huang]{guo2022recdis}
Yufei Guo, Xinyi Tong, Yuanpei Chen, Liwen Zhang, Xiaode Liu, Zhe Ma, and Xuhui Huang.
\newblock Recdis-snn: Rectifying membrane potential distribution for directly training spiking neural networks.
\newblock In \emph{Proceedings of the IEEE/CVF conference on computer vision and pattern recognition}, pages 326--335, 2022{\natexlab{b}}.

\bibitem[He et~al.(2015)He, Zhang, Ren, and Sun]{he2015delving}
Kaiming He, Xiangyu Zhang, Shaoqing Ren, and Jian Sun.
\newblock Delving deep into rectifiers: Surpassing human-level performance on imagenet classification.
\newblock In \emph{Proceedings of the IEEE international conference on computer vision}, pages 1026--1034, 2015.

\bibitem[Hinton et~al.(2015)Hinton, Vinyals, and Dean]{hinton2015distilling}
Geoffrey Hinton, Oriol Vinyals, and Jeff Dean.
\newblock Distilling the knowledge in a neural network.
\newblock \emph{arXiv preprint arXiv:1503.02531}, 2015.

\bibitem[Hu et~al.(2024)Hu, Deng, Wu, Yao, and Li]{hu2024advancing}
Yifan Hu, Lei Deng, Yujie Wu, Man Yao, and Guoqi Li.
\newblock Advancing spiking neural networks toward deep residual learning.
\newblock \emph{IEEE Transactions on Neural Networks and Learning Systems}, 2024.

\bibitem[IEA(2024)]{Iea}
IEA.
\newblock Electricity 2024.
\newblock https://www.iea.org/reports/electricity-2024, 2024.
\newblock Paris.

\bibitem[Jang et~al.(2021)Jang, Skatchkovsky, and Simeone]{jang2021bisnn}
Hyeryung Jang, Nicolas Skatchkovsky, and Osvaldo Simeone.
\newblock Bisnn: training spiking neural networks with binary weights via bayesian learning.
\newblock In \emph{2021 IEEE Data Science and Learning Workshop (DSLW)}, pages 1--6. IEEE, 2021.

\bibitem[Kheradpisheh et~al.(2022)Kheradpisheh, Mirsadeghi, and Masquelier]{kheradpisheh2022bs4nn}
Saeed~Reza Kheradpisheh, Maryam Mirsadeghi, and Timoth{\'e}e Masquelier.
\newblock Bs4nn: Binarized spiking neural networks with temporal coding and learning.
\newblock \emph{Neural Processing Letters}, 54\penalty0 (2):\penalty0 1255--1273, 2022.

\bibitem[Kingma and Ba(2014)]{kingma2014adam}
Diederik~P Kingma and Jimmy Ba.
\newblock Adam: A method for stochastic optimization.
\newblock \emph{arXiv preprint arXiv:1412.6980}, 2014.

\bibitem[Leroux et~al.(2020)Leroux, Vankeirsbilck, Verbelen, Simoens, and Dhoedt]{leroux2020training}
Sam Leroux, Bert Vankeirsbilck, Tim Verbelen, Pieter Simoens, and Bart Dhoedt.
\newblock Training binary neural networks with knowledge transfer.
\newblock \emph{Neurocomputing}, 396:\penalty0 534--541, 2020.

\bibitem[Li et~al.(2016)Li, Liu, Wang, Zhang, and Yan]{li2016ternary}
Fengfu Li, Bin Liu, Xiaoxing Wang, Bo~Zhang, and Junchi Yan.
\newblock Ternary weight networks.
\newblock \emph{arXiv preprint arXiv:1605.04711}, 2016.

\bibitem[Li et~al.(2017)Li, Liu, Ji, Li, and Shi]{li2017CIFAR10}
Hongmin Li, Hanchao Liu, Xiangyang Ji, Guoqi Li, and Luping Shi.
\newblock Cifar10-dvs: an event-stream dataset for object classification.
\newblock \emph{Frontiers in neuroscience}, 11:\penalty0 309, 2017.

\bibitem[Li et~al.(2024)Li, Geller, Kim, and Panda]{li2024seenn}
Yuhang Li, Tamar Geller, Youngeun Kim, and Priyadarshini Panda.
\newblock Seenn: Towards temporal spiking early exit neural networks.
\newblock \emph{Advances in Neural Information Processing Systems}, 36, 2024.

\bibitem[Lin and Yuan(2020)]{lin2020scalable}
Jie Lin and Jiann-Shiun Yuan.
\newblock A scalable and reconfigurable in-memory architecture for ternary deep spiking neural network with reram based neurons.
\newblock \emph{Neurocomputing}, 375:\penalty0 102--112, 2020.

\bibitem[Lu and Sengupta(2020)]{lu2020exploring}
Sen Lu and Abhronil Sengupta.
\newblock Exploring the connection between binary and spiking neural networks.
\newblock \emph{Frontiers in neuroscience}, 14:\penalty0 535, 2020.

\bibitem[Maass(1997)]{maass1997networks}
Wolfgang Maass.
\newblock Networks of spiking neurons: the third generation of neural network models.
\newblock \emph{Neural networks}, 10\penalty0 (9):\penalty0 1659--1671, 1997.

\bibitem[Neftci et~al.(2019)Neftci, Mostafa, and Zenke]{neftci2019surrogate}
Emre~O Neftci, Hesham Mostafa, and Friedemann Zenke.
\newblock Surrogate gradient learning in spiking neural networks: Bringing the power of gradient-based optimization to spiking neural networks.
\newblock \emph{IEEE Signal Processing Magazine}, 36\penalty0 (6):\penalty0 51--63, 2019.

\bibitem[Nguyen et~al.(2022)Nguyen, Tran, Dang, and Iacopi]{nguyen2022low}
Duy-Anh Nguyen, Xuan-Tu Tran, Khanh~N Dang, and Francesca Iacopi.
\newblock A low-power, high-accuracy with fully on-chip ternary weight hardware architecture for deep spiking neural networks.
\newblock \emph{Microprocessors and Microsystems}, 90:\penalty0 104458, 2022.

\bibitem[Orchard et~al.(2021)Orchard, Frady, Rubin, Sanborn, Shrestha, Sommer, and Davies]{orchard2021efficient}
Garrick Orchard, E~Paxon Frady, Daniel Ben~Dayan Rubin, Sophia Sanborn, Sumit~Bam Shrestha, Friedrich~T Sommer, and Mike Davies.
\newblock Efficient neuromorphic signal processing with loihi 2.
\newblock In \emph{2021 IEEE Workshop on Signal Processing Systems (SiPS)}, pages 254--259. IEEE, 2021.

\bibitem[Paszke et~al.(2019)Paszke, Gross, Massa, Lerer, Bradbury, Chanan, Killeen, Lin, Gimelshein, Antiga, et~al.]{paszke2019pytorch}
Adam Paszke, Sam Gross, Francisco Massa, Adam Lerer, James Bradbury, Gregory Chanan, Trevor Killeen, Zeming Lin, Natalia Gimelshein, Luca Antiga, et~al.
\newblock Pytorch: An imperative style, high-performance deep learning library.
\newblock \emph{Advances in neural information processing systems}, 32, 2019.

\bibitem[Pei et~al.(2023)Pei, Xu, Wu, Liu, and Yang]{pei2023albsnn}
Yijian Pei, Changqing Xu, Zili Wu, Yi~Liu, and Yintang Yang.
\newblock Albsnn: ultra-low latency adaptive local binary spiking neural network with accuracy loss estimator.
\newblock \emph{Frontiers in Neuroscience}, 17:\penalty0 1225871, 2023.

\bibitem[Qiu et~al.(2024)Qiu, Ning, Song, Fang, Chen, Sun, Ma, Yuan, and Tian]{qiu2024self}
Haonan Qiu, Munan Ning, Zeyin Song, Wei Fang, Yanqi Chen, Tao Sun, Zhengyu Ma, Li~Yuan, and Yonghong Tian.
\newblock Self-architectural knowledge distillation for spiking neural networks.
\newblock \emph{Neural Networks}, page 106475, 2024.

\bibitem[Ramasinghe et~al.(2023)Ramasinghe, MacDonald, Farazi, Saratchandran, and Lucey]{ramasinghe2023much}
Sameera Ramasinghe, Lachlan~Ewen MacDonald, Moshiur Farazi, Hemanth Saratchandran, and Simon Lucey.
\newblock How much does initialization affect generalization?
\newblock In \emph{International Conference on Machine Learning}, pages 28637--28655. PMLR, 2023.

\bibitem[Razani et~al.(2021)Razani, Morin, Sari, and Nia]{razani2021adaptive}
Ryan Razani, Gr{\'e}goire Morin, Eyyub Sari, and Vahid~Partovi Nia.
\newblock Adaptive binary-ternary quantization.
\newblock In \emph{Proceedings of the IEEE/CVF conference on computer vision and pattern recognition}, pages 4613--4618, 2021.

\bibitem[Romero et~al.(2014)Romero, Ballas, Kahou, Chassang, Gatta, and Bengio]{romero2014fitnets}
Adriana Romero, Nicolas Ballas, Samira~Ebrahimi Kahou, Antoine Chassang, Carlo Gatta, and Yoshua Bengio.
\newblock Fitnets: Hints for thin deep nets.
\newblock \emph{arXiv preprint arXiv:1412.6550}, 2014.

\bibitem[Shrestha et~al.(2024)Shrestha, Timcheck, Frady, Campos-Macias, and Davies]{shrestha2024efficient}
Sumit~Bam Shrestha, Jonathan Timcheck, Paxon Frady, Leobardo Campos-Macias, and Mike Davies.
\newblock Efficient video and audio processing with loihi 2.
\newblock In \emph{ICASSP 2024-2024 IEEE International Conference on Acoustics, Speech and Signal Processing (ICASSP)}, pages 13481--13485. IEEE, 2024.

\bibitem[Srivastava et~al.(2014)Srivastava, Hinton, Krizhevsky, Sutskever, and Salakhutdinov]{srivastava2014dropout}
Nitish Srivastava, Geoffrey Hinton, Alex Krizhevsky, Ilya Sutskever, and Ruslan Salakhutdinov.
\newblock Dropout: a simple way to prevent neural networks from overfitting.
\newblock \emph{The journal of machine learning research}, 15\penalty0 (1):\penalty0 1929--1958, 2014.

\bibitem[Wan et~al.(2013)Wan, Zeiler, Zhang, Le~Cun, and Fergus]{wan2013regularization}
Li~Wan, Matthew Zeiler, Sixin Zhang, Yann Le~Cun, and Rob Fergus.
\newblock Regularization of neural networks using dropconnect.
\newblock In \emph{International conference on machine learning}, pages 1058--1066. PMLR, 2013.

\bibitem[Wu et~al.(2023)Wu, Chen, and Huang]{wu2023hardware}
Nai-Chun Wu, Tsu-Hsiang Chen, and Chih-Tsun Huang.
\newblock Hardware-aware model architecture for ternary spiking neural networks.
\newblock In \emph{2023 International VLSI Symposium on Technology, Systems and Applications (VLSI-TSA/VLSI-DAT)}, pages 1--4. IEEE, 2023.

\bibitem[Wu et~al.(2019)Wu, Deng, Li, Zhu, Xie, and Shi]{wu2019direct}
Yujie Wu, Lei Deng, Guoqi Li, Jun Zhu, Yuan Xie, and Luping Shi.
\newblock Direct training for spiking neural networks: Faster, larger, better.
\newblock In \emph{Proceedings of the AAAI conference on artificial intelligence}, volume~33, pages 1311--1318, 2019.

\bibitem[Xiao et~al.(2017)Xiao, Rasul, and Vollgraf]{xiao2017/online}
Han Xiao, Kashif Rasul, and Roland Vollgraf.
\newblock Fashion-mnist: a novel image dataset for benchmarking machine learning algorithms, 2017.

\bibitem[Xu et~al.(2023)Xu, Li, Shen, Liu, Tang, and Pan]{xu2023constructing}
Qi~Xu, Yaxin Li, Jiangrong Shen, Jian~K Liu, Huajin Tang, and Gang Pan.
\newblock Constructing deep spiking neural networks from artificial neural networks with knowledge distillation.
\newblock In \emph{Proceedings of the IEEE/CVF Conference on Computer Vision and Pattern Recognition}, pages 7886--7895, 2023.

\bibitem[Yang et~al.(2021)Yang, Martinez, Bulat, Tzimiropoulos, et~al.]{yang2021knowledge}
Jing Yang, Brais Martinez, Adrian Bulat, Georgios Tzimiropoulos, et~al.
\newblock Knowledge distillation via softmax regression representation learning.
\newblock International Conference on Learning Representations (ICLR), 2021.

\bibitem[Yousefzadeh et~al.(2022)Yousefzadeh, Van~Schaik, Tahghighi, Detterer, Traferro, Hijdra, Stuijt, Corradi, Sifalakis, and Konijnenburg]{yousefzadeh2022seneca}
Amirreza Yousefzadeh, Gert-Jan Van~Schaik, Mohammad Tahghighi, Paul Detterer, Stefano Traferro, Martijn Hijdra, Jan Stuijt, Federico Corradi, Manolis Sifalakis, and Mario Konijnenburg.
\newblock Seneca: Scalable energy-efficient neuromorphic computer architecture.
\newblock In \emph{2022 IEEE 4th International Conference on Artificial Intelligence Circuits and Systems (AICAS)}, pages 371--374. IEEE, 2022.

\bibitem[Yun et~al.(2019)Yun, Han, Oh, Chun, Choe, and Yoo]{yun2019cutmix}
Sangdoo Yun, Dongyoon Han, Seong~Joon Oh, Sanghyuk Chun, Junsuk Choe, and Youngjoon Yoo.
\newblock Cutmix: Regularization strategy to train strong classifiers with localizable features.
\newblock In \emph{Proceedings of the IEEE/CVF international conference on computer vision}, pages 6023--6032, 2019.

\bibitem[Zhang et~al.(2018)Zhang, Cisse, Dauphin, and Lopez-Paz]{zhang2018mixup}
Hongyi Zhang, Moustapha Cisse, Yann~N. Dauphin, and David Lopez-Paz.
\newblock mixup: Beyond empirical risk minimization.
\newblock In \emph{International Conference on Learning Representations}, 2018.
\newblock URL \url{https://openreview.net/forum?id=r1Ddp1-Rb}.

\bibitem[Zhong et~al.(2020)Zhong, Zheng, Kang, Li, and Yang]{zhong2020random}
Zhun Zhong, Liang Zheng, Guoliang Kang, Shaozi Li, and Yi~Yang.
\newblock Random erasing data augmentation.
\newblock In \emph{Proceedings of the AAAI conference on artificial intelligence}, volume~34, pages 13001--13008, 2020.

\bibitem[Zhou et~al.(2022)Zhou, Lei, Liu, Du, Huang, Zhao, Dai, Chen, Le, and Laudon]{NEURIPS2022_2f00ecd7}
Yanqi Zhou, Tao Lei, Hanxiao Liu, Nan Du, Yanping Huang, Vincent Zhao, Andrew~M Dai, zhifeng Chen, Quoc~V Le, and James Laudon.
\newblock Mixture-of-experts with expert choice routing.
\newblock In S.~Koyejo, S.~Mohamed, A.~Agarwal, D.~Belgrave, K.~Cho, and A.~Oh, editors, \emph{Advances in Neural Information Processing Systems}, volume~35, pages 7103--7114. Curran Associates, Inc., 2022.
\newblock URL \url{https://proceedings.neurips.cc/paper_files/paper/2022/file/2f00ecd787b432c1d36f3de9800728eb-Paper-Conference.pdf}.

\bibitem[Zhou et~al.(2023)Zhou, Zhu, He, Wang, YAN, Tian, and Yuan]{zhou2023spikformer}
Zhaokun Zhou, Yuesheng Zhu, Chao He, Yaowei Wang, Shuicheng YAN, Yonghong Tian, and Li~Yuan.
\newblock Spikformer: When spiking neural network meets transformer.
\newblock In \emph{The Eleventh International Conference on Learning Representations}, 2023.
\newblock URL \url{https://openreview.net/forum?id=frE4fUwz_h}.

\bibitem[Zhu et~al.(2023)Zhu, Zhao, and Eshraghian]{zhu2023spikegpt}
Rui-Jie Zhu, Qihang Zhao, and Jason~K Eshraghian.
\newblock Spikegpt: Generative pre-trained language model with spiking neural networks.
\newblock \emph{arXiv preprint arXiv:2302.13939}, 2023.

\bibitem[Zhu et~al.(2024)Zhu, Zhang, Zhao, Deng, Duan, and Deng]{zhu2024tcja}
Rui-Jie Zhu, Malu Zhang, Qihang Zhao, Haoyu Deng, Yule Duan, and Liang-Jian Deng.
\newblock Tcja-snn: Temporal-channel joint attention for spiking neural networks.
\newblock \emph{IEEE Transactions on Neural Networks and Learning Systems}, 2024.

\end{thebibliography}

\end{document}